\documentclass{article}

\usepackage[preprint]{neurips_2019}

\usepackage[utf8]{inputenc} % allow utf-8 input
\usepackage[T1]{fontenc}    % use 8-bit T1 fonts
\usepackage[colorlinks]{hyperref}       % hyperlinks
\usepackage{url}            % simple URL typesetting
\usepackage{booktabs}       % professional-quality tables
\usepackage{amsfonts}       % blackboard math symbols
\usepackage{nicefrac}       % compact symbols for 1/2, etc.
\usepackage{microtype}      % microtypography
\usepackage{amsmath,amssymb,amsthm}
\usepackage{verbatim,float,url,enumerate}
\usepackage{graphicx,subfigure,psfrag}
\usepackage{natbib,xcolor}
\usepackage{microtype}
\usepackage{tikz}
\usetikzlibrary{bayesnet}
\usetikzlibrary{arrows}
\usetikzlibrary{backgrounds,positioning}
\usetikzlibrary{decorations.pathreplacing}
\usepackage{caption}
\usepackage{wrapfig,lipsum,booktabs}

\newtheorem{theorem}{Theorem}
\newtheorem{lemma}{Lemma}

\theoremstyle{remark}

\theoremstyle{definition}

\newcommand{\argmin}{\mathop{\mathrm{argmin}}}
\newcommand{\argmax}{\mathop{\mathrm{argmax}}}

\newcommand\ci{\perp\!\!\!\perp}

\def\R{\mathbb{R}}
\def\E{\mathbb{E}}
\def\P{\mathbb{P}}

\def\rva{\alpha_{z_i}}
\def\rvb{\eta_{z_{-i}}}
\def\lrva{\mathcal{L}(\alpha_{z_i})}
\def\lrvb{\mathcal{L}(\eta_{z_{-i}})}
\def\urva{\mathcal{U}(\alpha_{z_i})}
\def\urvb{\mathcal{U}(\eta_{z_{-i}})}
\title{Better Approximate Inference for Partial Likelihood Models with a Latent Structure}

\author{%
  Amrith Setlur\\
  School of Computer Science\\
  Carnegie Mellon University\\
  Pittsburgh, PA 15213 \\
  \texttt{asetlur@cs.cmu.edu} \\
  \And
  Barnab\'as  P\'ocz\'os\\
  School of Computer Science\\
  Carnegie Mellon University\\
  Pittsburgh, PA 15213 \\
  \texttt{bapoczos@cs.cmu.edu} \\
}

\begin{document}

\maketitle

\begin{abstract}

Temporal Point Processes (TPP) with partial likelihoods involving a latent structure often entail an intractable marginalization, thus making inference hard. We propose a novel approach to Maximum Likelihood Estimation (MLE) involving approximate inference over the latent variables by minimizing a tight upper bound on the approximation gap. Given a discrete latent variable $Z$, the proposed approximation reduces inference complexity from $\mathcal{O}(|Z|^c)$ to $\mathcal{O}(|Z|)$. We use convex conjugates to determine this upper bound in a closed form and show that its addition to the optimization objective results in improved results for models assuming proportional hazards as in Survival Analysis. %Our inference strategy is applicable to the full likelihood in a TPP, we focus on its impact in the case of partial likelihoods since the objective there is closely related to the MLE objectives observed in latent Conditional Random Fields (CRFs), thus making our work applicable to a broader class of problems. %Finally, we show that optimizing the combined objective improves the concordance index in the Cox Proportional Hazards (CPH) model for Survival Analysis. %while the approximation bound is We also provide a tight closed-form upper bound on our approximation which can be exploited in the overall optimization objective. 
%Statistical tools like Gibbs sampling \& Monte Carlo based methods tailored for specific TPP models are not only computationally expensive but they may also fail to converge. 
\end{abstract}

\section{Introduction}
\label{sec:introduction}

Temporal Point Processes (TPPs) provide a formal framework to model the occurrences of discrete events in time (like failures or financial transactions). Recent work (\citep{linderman2014discovering} \citep{snoek2013determinantal}) on modelling TPPs with latent factors have showcased their ability to capture correlations such as inhibitory relationships \& a dichotomy of classes of neurons in neural spike recordings. Although, there have been several advances in non-parametric Bayesian inference (\citep{samo2015scalable}) most models are parametric (\citep{cox1955some}) where parameter estimation is done by maximizing the likelihood of observed point values.
 %Latent variables are commonly used to model causality in biological/psychological Probabilistic Graphical Models (PGMs) or to regress on unobserved factors in economic trends \citep{koller2009probabilistic}. 
  %Models in \citep{snoek2013determinantal} \citep{klein2011mixture} have relied on stochastic methods like Markov Chain Monte Carlo (MCMC) for inference. \citep{cowles1996markov} highlight the intricacies of diagnosing the convergence of computationally expensive MCMC methods which are often custom-tailored. 
  Survival analysis is the problem of estimating survival times for entities (like nodes in a machine) and it has largely relied on TPPs to estimate survival times in the presence of censored observations. Semi-parametric methods like the Cox Proportional Hazards (CPH) \citep{cox1955some} allow parametric estimation using a partial likelihood objective without estimating the baseline hazard.  Therefore, we propose an approximate inference strategy for latent variable models with a partial likelihood objective.  We introduce an inference method for models where the normalization factor includes interactions over log-linear factors. Such models are common in TPPs assuming proportional hazards (\citep{rosen1999mixtures}) or in latent Conditional Random Fields (CRFs) where the normalization involves a sum over finite potential functions (\citep{sutton2012introduction}). \citep{rosen1999mixtures} introduce an inference strategy for CPH which is similar to our proposed method, but they fail to identify cases where the approximation fails. Although our inference strategy is applicable to the full likelihood in a TPP, we focus on its impact in the case of partial likelihoods since the objective there is closely related to the MLE objectives observed in latent CRFs, thus making our work applicable to a broader class of problems.
 
  %have been the one of the most successful to estimate parameters class of  for This particular class of TPPs warrants the addition of latent variables to model  involving parameter estimation in a TPP, wherein certain observations are censored 
% Therefore, we propose an approximate inference strategy for latent variable models with a partial likelihood objective.  We introduce an inference method for models where the normalization factor includes interactions over log-linear factors. Such models are common in partial-likelihood problems (\citep{rosen1999mixtures}) or latent CRFs where the normalization involves a sum over finite potential functions (\citep{sutton2012introduction}). Although \citep{rosen1999mixtures} introduce an inference strategy for Cox Proportional Hazards (CPH) which is similar to our proposed method, they fail to identify cases where the approximation fails. Our inference strategy is applicable to the full likelihood in a TPP, we focus on its impact in the case of partial likelihoods since the objective there is closely related to the MLE objectives observed in latent Conditional Random Fields (CRFs), thus making our work applicable to a broader class of problems.
 
 Inspired by \citep{jebara2012majorization} we introduce a distribution agnostic closed form tight upper bound on likelihood estimations for TPPs resembling \citep{diggle2005partial}. The upper bound can be minimized via standard gradient descent based iterative methods \citep{ruder2016overview}. Finally, we prove a tight upper bound on the Jensen inequality for strictly convex polynomial functions on $\R_{++}$.
 %Poisson Point Process are closely related to models on survival analysis as both of them have identical definitions for the hazard function\citep{fisher1999time}, where the risk of an event is independent of its history. Inference methods in TPPs with Cox/Poisson Process Models or variational inference in PGMs often involve a trade-off between computational efficiency and  stochastic accuracy. 
 %gap via a closed form function which can optimized via standard gradient descent based iterative methods \citep{ruder2016overview}.
  % These are latent variable mixture models with PGMs similar to figure \ref{fig:g_model} whose objective can be characterised by the adjacent equation \ref{fig:g_model}.  
 %Furthermore, we introduce a tight upper bound on the approximation gap (between the true MLE and our approximation) in a closed form which can be reduced by Moreover we model the approximation gap via a closed form function which can optimized via standard gradient descent based iterative methods \citep{ruder2016overview}.
\section{The Inference Problem}
\label{sec:model}

Given a compact set $S$ equipped with Borel $\sigma$-algebra $\mathbb{B}(S)$, $X(t):\Omega\to\Gamma$ is a TPP if $X(t)$ is a measurable transform from the probability space $(\Omega, \mathfrak{F}, \P)$ to the space of counting measures $\Gamma$ on S. Given a series of events $\{(\delta_i, x_i, t_i)\}_N$: for a given entity $x_i$ the event of interest occurred $(\delta_i=1)$ at time $t_i$ or the observation was censored $(\delta_i=0)$. The risk set for $x_i$ is given by $R(t_i) = \{x_j : t_j \geq t_i\}$. Under a Poisson TPP with intensity $\lambda(x)$, the likelihood of the event $\delta_i=1$ for $x_i$ at $t_i$ given that the event hasn't occurred till $t_i$ is given by $\P(N(B_a)=1)/\P(N(B_b)=0)$. $B_a,B_b \in \mathbb{B}, B_a=[t_i, t_i+\delta), B_b=[0, t_i)$ where $N(B)$ follows a Poisson distribution with mean $\int_{B}\lambda(x)dx$. 

We modify the formulation by adding latent variables $z$ (see figure \ref{fig:g_model}) and now the intensity function in the TPP is function of parameters $\beta$, input $x_i$ and latent variable $z_i \sim p(.|x_i,\theta)$. \citep{rosen1999mixtures}  and \citep{diggle2005partial} used partial likelihood models to efficiently compute the MLE estimates for the parameters in an inhomogenous Poisson process. Partial likelihood was first introduced by \citep{cox1955some} with the aim of identifying variables that impact survival analysis without worrying about the baseline hazard. For the same reasons, we choose to maximize the partial likelihood of an event $(\delta_i, x_i, t_i)$  conditioned on the risk set $R(t_i)$. Thus the denominator in eq. \ref{fig:g_model} now involves a sum over a finite set of factors from $R(t_i)$ (closely resembling latent CRFs \citep{quattoni2007hidden} in likelihood estimation). 
\vspace{-2em}
\begin{align}
  \vcenter{\hbox{
  \begin{tikzpicture}[thick,scale=0.7, every node/.style={scale=0.7}, every plate/.style={scale=0.7}, every edge/.style={scale=0.7}]
  \label{fig:g_model}
  \begin{scope}
    % nodes
      \node[latent] (lambda) {$\lambda(t)$}; %
     \node[obs, right=of lambda,inner sep=0pt] (x) {$x$};%
     \node[latent,below=of lambda] (z) {$z$}; %
      \node[latent,left=of lambda] (beta) {$\beta$};
      \node[latent,left=of z] (theta) {$\theta$};
    % plate
     \plate[inner sep=.87cm, yshift=0.17cm]{plate1}{(x)(z)(lambda)}{$N$}; %
    \end{scope}
    % edges
     \edge {x} {z}  
     \edge {x} {lambda}  
     \edge {beta} {lambda}  
     \edge {theta} {z}
     \edge {z} {lambda}
 \end{tikzpicture}}} \quad\quad
    \P(\delta_i|x_i, R(t_i), \beta) &= \E_{z_i \sim p(.| \theta, x_i)} \left[ \frac{\exp{\beta_{z_i}^Tx_i}}{\sum\limits_{j \in R(t_i)} \exp{\beta_{z_j}^Tx_j}}\right]
\end{align}
%One way to alleviate this is by proposing a partial likelihood maximization as done by \citep{rosen1999mixtures} or \citep{diggle2005partial}. 
\section{Approximate Inference Solution}
\label{sec:approx inference}
In this section we provide a computationally tractable approximation for the maximum-likelihood estimation of the semi-parametric latent variable model defined in section \ref{sec:model} and in section \ref{sec:analysis} we show the conditions under which the approximation is tight. Assuming $z_i \ci z_{j, j\neq i} \, | \, x_i,x_j$ we can define positive random variables (R.V.) $\rva$ and $\rvb$ which are functions of $\beta, R(t_i)$ \& $z$. This assumption implies that $\rva \ci \rvb \, | \, x_i,x_j$. In the rest of the paper (unless stated otherwise), the expectation $\E$ is over the distribution $z_i \sim p(z_i| \theta, x_i)$. Using this re-formulation and the Taylor series expansion we can re-write eq. \ref{fig:g_model} as,
\vspace{-1em}
 \begin{align}
  \label{eq:def_ab_1}
 \rva = \exp{\beta_{z_i}^Tx_i}  \quad  \rvb=\sum\limits_{j \in R(t_i)} \exp{\beta_{z_j}^Tx_j} \quad P(\delta_i|x_i, R(t_i)) = \E \left( \frac{\rva+\rvb}{\rva} \right)^{-1}
 \end{align}
 \vspace{-1.5em}
 \begin{align}
    \label{eq:approx}
    \E \left[\left(\frac{\rva+\rvb}{\rva} \right)^{-1}\right] = \E\left[\sum\limits_{p=1}^\infty  (-1)^p \E(\rvb^p)\E(\rva^{-p})\right] \approx \left[\sum\limits_{p=1}^K (-1)^p \E(\rvb^p)\E(\rva^{-p})\right]
\end{align}
 \begin{lemma}
 \label{lemma:moment}
 If we assume $\rva$, $\rvb$ to have moments of order $\mathcal{H}_{\rva}, \mathcal{H}_{\rvb}$ respectively, then their ratio distribution will have moments of order $\frac{\mathcal{H}_{\rva}\mathcal{H}_{\rvb}}{\mathcal{H}_{\rva}+\mathcal{H}_{\rvb}}$. \citep{cedilnik2006ratio}
 \end{lemma}
Using the Mellin Transform theory for ratio distributions of positive independent random variables (R.V.); we have $\E((X/Y)^p) = \E(X^p)\E(Y^{-p})$.  Based on lemma \ref{lemma:moment} we limit the expansion in eq. \ref{eq:approx} to a finite $K$.
 At this point, we are computing expectations over convex functions $x^p$ and $x^{-p}$ with $p>0$ (defined on $\R_{++}$). Since for a convex $f$, $E(f(x))\geq f(E(x))$ (Jensen inequality) we can further approximate eq. \ref{eq:approx} with eq. \ref{eq:approx2}. Once again we use the Taylor series approximation to finally arrive at a tractable maximum-likelihood objective (eq. \ref{eq:approx3}). For each data point the inference complexity under the original objective is $\mathcal{O}(|z|^{|R(t_i)|})$ whereas under the proposed marginalization the complexity reduces to $\mathcal{O}(|z|)$.
 \vspace{-1em}
\begin{align}
    \label{eq:approx2}
    \E \left[\left(\frac{\rva+\rvb}{\rva} \right)^{-1}\right] \approx \sum\limits_{p=1}^K (\E(\rvb))^p(\E(\rva))^{-p} \quad
    \P(\delta_i|x_i, R(t_i), \beta) &= \left[ \frac{\E(\exp{\beta_{z_i}^Tx_i})}{\sum\limits_{j \in R(t_i)} \E(\exp{\beta_{z_j}^Tx_j})}\right]
\end{align}

 %define the moment generating functions for $\rva$ \& $\rvb$ respectively, then the assumption that $M_{\rva}$ (and $M_{\rvb}$) is finite in an open interval containing $0$ is not a strong one. 

\begin{align}
    \label{eq:approx3}
    \P(\delta_i|x_i, R(t_i), \beta) &= \left[ \frac{\E_{z_i \sim p(.| \theta, x_i)}(\exp{\beta_{z_i}^Tx_i})}{\sum\limits_{j \in R(t_i)} \E_{z_j \sim p(.| \theta, x_j)}(\exp{\beta_{z_j}^Tx_j})}\right]
\end{align}

The crux of the approximation lies in the Jensen inequality. Therefore, we spend the following section on identifying a tight distribution independent bound on the inequality gap. If this inequality gap is reduced then we know that \ref{eq:approx3} is a good approximation for \ref{fig:g_model}.
 
% \begin{align} 
% \label{eq:def_ab_exp}
% \E_{z_i \sim p(.| \theta, x_i)} \left[\left(\frac{\rva+\rvb}{\rva} \right)^{-1}\right] = 1 + \frac{\rvb}{\rva} + \frac{\rvb^2}{\rva^2} + \dots + (-1)^k \, \frac{\rvb^k}{\rva^k} + \dots  
% \end{align}
\vspace{-1em}
\section{Bounding the Approximation [Analysis]}
\label{sec:analysis}
We identify the conditions under which the approximation \ref{eq:approx} is feasible and provide a closed form bound for it. In order to simplify the statements in the rest of the paper, we introduce some notations and assumptions here. We assume that the R.V. $z_i \sim p_\theta(.|x_i)$ has mean $\mu(x_i)$ and $z_i-\mu(x_i)$ is sub-Gaussian with parameter $\sigma(x_i)$. This assumption is fairly common in latent variable models where the true posterior $p(z_i|x_i)$ is approximated by a Gaussian distribution $N(\mu(x_i), \sigma(x_i))$. For a continuous function $\beta_{z_i} = \beta(z_i)$, with $z_i$ lying in a closed, bounded set (with probability $\delta$), one can bound the values attained by $\exp{\beta_{z_i}^Tx_i}$. Thus for a R.V. $z_i$ we obtain reasonable probabilistic bounds on $\rva, \rvb$, formalized by the following statement: $\rva \in [\lrva, \urva], \rvb \in [\lrvb, \urvb]$ with probability $\delta$ (defined in theorem \ref{theorem:main})).

\begin{theorem}
    \label{theorem:main}
    Although this is stated for $\rva$, the statement for $\rvb$ is similar. \\
    With probability  $\delta=erf\left(\frac{1}{2}\frac{\urva-\exp{\beta_{z_i}^T\mu(x_i)}}{\sqrt{2}\sigma(x_i)(\exp{\beta_{z_i}^T\mu(x_i)})\|\beta_{z_i}\|_2}\right)-erf\left(\frac{1}{2}\frac{-\lrva-\exp{\beta_{z_i}^T\mu(x_i)}}{\sqrt{2}\sigma(x_i)(\exp{\beta_{z_i}^T\mu(x_i)})\|\beta_{z_i}\|_2}\right)$,
    \begin{align}
        \label{eq:theorem_equation}
             \E(\phi_p(\rva)) - \phi_p(E(\rva)) &\leq \kappa\phi_p(\urva) + (1-\kappa)\phi_p(\lrva) - \phi_p(\kappa\urva + (1-\kappa)\lrva) \\
            \kappa &= \nabla\phi*\left[\frac{\phi_p(\urva)-\phi_p(\lrva)}{\urva-\lrva}\right]
    \end{align}
    The conjugate $\phi_p^*(y) = \sup\limits_{x} y^Tx-\phi_p(x)$, $\nabla \phi_p^*(y) = \{x : y^Tx-\phi_p(x)=\phi_p^*(y)\}$, which is a singleton set for strictly convex functions.
\end{theorem}
\vspace{-1em}
\section{Joint Objective Function}
\vspace{-1.2em}
Since gradients for conjugate functions $\phi_p^*(x)$ are well defined in our case (see Appendix \ref{appendix:tight_bound}), we can show that the approximation in eq. \ref{eq:approx3} is good when the joint objective is minimized (eq. \ref{eq:new_objective}). The joint objective enforces the model to find optimal $(\theta^*, \beta^*)$ that maximizes the likelihood in eq. \ref{fig:g_model}, while ensuring proximity to the true objective. Eq. \ref{eq:new_objective} can also be viewed from the perspective of a regularized objective where the model learns to enforce additional constraints on the variance of $\rva, \rvb$ and thus ends up with distributions of $z_i$ with rapidly decaying Gaussian tails. %In section \ref{sec:results} we show results on CPH mixture models for survival analysis where the regularization is implicitly enforced by gating the predictions and ensuring peaky distributions of the discrete latent R.V. $z_i$. 
\begin{align}
    \label{eq:new_objective}
        (\beta^*, \theta^*) = \argmin_{\beta, \theta} \sum_{i=1}^{N}\left[ \frac{\E_{z_i \sim p(.| \theta, x_i)}(\exp{\beta_{z_i}^Tx_i})}{\sum\limits_{j \in R(t_i)} \E_{z_j \sim p(.| \theta, x_j)}(\exp{\beta_{z_j}^Tx_j})} + \sum\limits_{j \in R(t_i)} L(x_j, \kappa_j, \theta, \beta)\right]
\end{align}
Here $L(x_j, \kappa_j, \theta, \beta)$ is obtained by using theorem \ref{theorem:main} which bounds the Jensen's inequality for each data point $(\delta_i, x_i, t_i)$, via a sum over gradients computed for the functions $\phi_p^*$.

% By Jensen's inequality and the convexity of $f(x)=x^p$, $f(x)=x^{-p}$ ($p>=1$); we have that $\E[\rvb^p] >= [\E(\rvb)]^p$. Similarly $\E[\rva^{-p}] >= [\E(\rva)]^{-p}$. Under certain conditions, the above gaps are much smaller in which case we can replace $\E[\rvb^{-p}]$, $\E[\rva^{-p}]$ with their corresponding lower bounds in eq. \ref{eq:approx}. In the following section we investigate this gap and try to quantify it via a closed form solution. This further facilitates the possibility of adding the closed form convexity gap as a part of the objective in the maximum-likelihood formulation (eq. \ref{fig:g_model}). 

% \subsection{Case of continuous latent variables}

% As suggested by \citep{doersch2016tutorial}, for optimization purposes, most models assume $z_i \sim p_{\theta}(.|x_i)$ to be a gaussian R.V. with a learnable mean $\mu_{\theta}({x_i})$ \& variance $\sigma^2_{\theta}(x_i)$. During parameter estimation, at each iteration we identify , 

\section{Results}
\vspace{-1.2em}
\label{sec:results}

We analyze two types of results: (1) we evaluate our combined objective (eq. \ref{eq:new_objective}) on a proportional hazards (CPH) model and show an improvement in the concordance-index (table \ref{tab:results1}), (2) we compare our proposed distribution agnostic bound against a standard bound for the Jensen inequality \citep{dragomir1999converse}.
\stepcounter{figure}
\begin{figure}
\centering
\begin{minipage}{0.48\textwidth}
 \centering
  \includegraphics[width=0.95\linewidth, height=4cm]{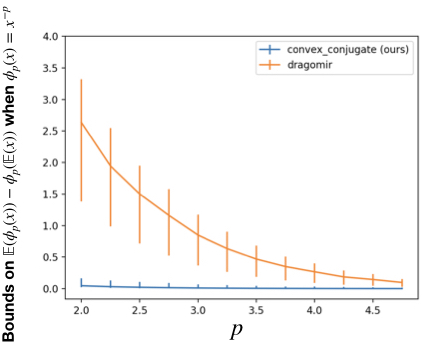}
  \captionof{figure}{Comparisons of the upper bounds on the Jensen inequality established by our model and the baseline \citep{dragomir1999converse}. 95\% confidence intervals are shown when values of $\urva\sim N(\mu_{u},\sigma^2)$ and $\lrva\sim N(\mu_{l},\sigma^2)$.}
  \label{fig:comparison_bound}
\end{minipage}\hfill
\begin{minipage}{0.48\textwidth}
  \centering
  \includegraphics[width=0.95\linewidth]{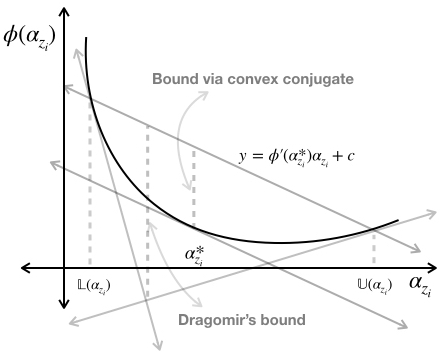}
  \captionof{figure}{Visualization of our distribution agnostic bound on Jensen's inequality for strictly convex functions.}
  \label{fig:vis_bound}
\end{minipage}
\end{figure}
\vspace{-1em}
\subsection{Survival Analysis}
\vspace{-1em}
Given a discrete $z$ we model the distribution $z_i \sim p(,|\theta, x_i)$ to be a multinomial. The final layer of the input encoder network is a softmax operation ensuring that the  distribution over the latent space of $z$ is a valid one. We compare our models: Latent Variable CPH via Hard/Soft Gating (LV-CPH-HG/LV-CPH-SG) against three popular baselines: \textsc{Cph} \citep{cox1955some}, \textsc{Rsf} \citep{ishwaran2007random}, \textsc{DeepSurv} \citep{katzman2016deep} on common datasets in survival analysis: \textsc{METABRIC} \citep{yao2014methods}, \textsc{ROTTERDAM-GBSG}  \citep{schumacher1994randomized}, \textsc{SUPPORT} \citep{knaus1995support}.
%We evaluate our proposed objective on CPH models for survival analysis. Here, we consider a mixture model where the latent variable is discrete, i.e. $z_i$ can take $1$ of $K$ values.

For the discrete case, it is easy to see that the regularizer $L(x_i, \kappa_i, \theta, \beta)$ in eq. \ref{eq:new_objective} is minimized when $z_i$ has low variance (or entropy). We enforce a low entropy distribution by gating (soft/hard) the predictions $p(.|\theta, x_i)$ obtained from the softmax layer. Since for the discrete case, low entropy ($\mathbb{H}_{z_i}$) on $z_i$ $\Longleftrightarrow$ $|\lrva -\urva| + |\lrvb - \urvb| < C$ and $C\to0$ as $\mathbb{H}_{z_i}\to0$, one can instead minimize $\mathbb{H}_{z_i}$ to effectively reduce the upper bound in theorem \ref{theorem:main}. Therefore, we conclude that optimizing for the joint objective function in eq. \ref{eq:new_objective} instead of the mere approximation in eq. \ref{eq:approx3} leads to an improved concordance-index for CPH models.
\vspace{-0.6em}
\subsection{Tightness of the proposed bound}
\vspace{-0.6em}
Figure \ref{fig:comparison_bound} compares the bound computed by \citep{dragomir1999converse} [baseline] (Appendix \ref{appendix:loose_bound}) against our tight bound (Appendix $\ref{appendix:tight_bound}$), by sampling $\urva, \urvb, \lrva, \lrvb$ from distinct fixed normal distributions. Our bound is much tighter for smaller values of $p$, and it converges to the baseline's value for large p. Looking at figure \ref{fig:vis_bound} it is easy to verify that the bound we propose on Jensen's inequality is stronger than the baseline.
\vspace{-0.em}
\begin{table}[!htbp]
  \centering
\begin{tabular}{|c|c|c|c|}\toprule
\textsc{ Model} &  \textsc{Metabric} &\textsc{ Rotterdam-Gbsg} &\textsc{ Support} \\ \hline
\textsc{Cph} & $0.6306\pm0.004$ & $0.6578\pm0.004$ & $0.5828\pm0.002$\\ 
\textsc{DeepSurv} & $0.6434\pm0.004$ & 0.$6684\pm0.003$ & $0.6183\pm0.002$ \\ 
\textsc{Rsf} & $0.6243\pm0.004$ & $0.6512\pm0.003$ & $0.6130\pm0.002$\\ \midrule
\textsc{LV-CPH-SG} & $\mathbf{0.6585\pm0.003}$ & $0.6752\pm0.002$ & $\mathbf{0.6196\pm0.001}$ \\
\textsc{LV-CPH-HG} & $0.6349\pm0.003$ & $\mathbf{0.6866\pm0.002}$ & $0.5706\pm0.001$ \\ \bottomrule
  \end{tabular}
  \vspace{0.5em}
   \caption{Results of hard and soft linear gating networks and their comparison with relevant baselines (95\% bootstrap CI ). 
   \vspace{-1.3em}
   \citep{nagpal2019nonlinear}}
     \label{tab:results1}
\vspace{-2.5mm}
\end{table}
\section{Discussion}
\vspace{-0.6em}
We propose an approximation for the MLE objective in TPPs involving a partial likelihood function with latent factors. We also show that the MLE approximation can be bounded by minimizing a joint objective which includes an upper bound on the approximation gap. We have shown this to be theoretically and empirically better for the partial likelihood estimation (in survival analysis). Future work on this inference method would be to further exploit the tractable closed form approximation gap by directly optimizing for it with iterative methods like ADAM. Yet another direction would be to extend this work to latent variable models for semi-parametric models like Gaussian processes for survival analysis \citep{fernandez2016gaussian}.
\small
\bibliographystyle{plainnat}
\bibliography{refs}

\begin{thebibliography}{20}
\providecommand{\natexlab}[1]{#1}
\providecommand{\url}[1]{\texttt{#1}}
\expandafter\ifx\csname urlstyle\endcsname\relax
  \providecommand{\doi}[1]{doi: #1}\else
  \providecommand{\doi}{doi: \begingroup \urlstyle{rm}\Url}\fi

\bibitem[Cedilnik et~al.(2006)Cedilnik, Kosmelj, and Blejec]{cedilnik2006ratio}
Anton Cedilnik, Katarina Kosmelj, and Andrej Blejec.
\newblock Ratio of two random variables: A note on the existence of its
  moments.
\newblock \emph{Metodoloski zvezki}, 3\penalty0 (1):\penalty0 1, 2006.

\bibitem[Cox(1955)]{cox1955some}
David~R Cox.
\newblock Some statistical methods connected with series of events.
\newblock \emph{Journal of the Royal Statistical Society: Series B
  (Methodological)}, 17\penalty0 (2):\penalty0 129--157, 1955.

\bibitem[Diggle(2005)]{diggle2005partial}
Peter~J Diggle.
\newblock A partial likelihood for spatio-temporal point processes.
\newblock \emph{Biostats}, 2005.

\bibitem[Dragomir(1999)]{dragomir1999converse}
Silvestru~S Dragomir.
\newblock A converse result for jensen’s discrete inequality via gr{\"u}ss’
  inequality and applications in information theory.
\newblock \emph{An. Univ. Oradea Fasc. Mat}, 1999.

\bibitem[Fern{\'a}ndez et~al.(2016)Fern{\'a}ndez, Rivera, and
  Teh]{fernandez2016gaussian}
Tamara Fern{\'a}ndez, Nicol{\'a}s Rivera, and Yee~Whye Teh.
\newblock Gaussian processes for survival analysis.
\newblock In \emph{Advances in Neural Information Processing Systems}, pages
  5021--5029, 2016.

\bibitem[Ishwaran and Lu(2007)]{ishwaran2007random}
Hemant Ishwaran and Min Lu.
\newblock Random survival forests.
\newblock \emph{Wiley StatsRef: Statistics Reference Online}, pages 1--13,
  2007.

\bibitem[Jebara and Choromanska(2012)]{jebara2012majorization}
Tony Jebara and Anna Choromanska.
\newblock Majorization for crfs and latent likelihoods.
\newblock In \emph{Advances in Neural Information Processing Systems}, pages
  557--565, 2012.

\bibitem[Katzman et~al.(2016)Katzman, Shaham, Cloninger, Bates, Jiang, and
  Kluger]{katzman2016deep}
Jared~L Katzman, Uri Shaham, Alexander Cloninger, Jonathan Bates, Tingting
  Jiang, and Yuval Kluger.
\newblock Deep survival: A deep cox proportional hazards network.
\newblock \emph{stat}, 1050:\penalty0 2, 2016.

\bibitem[Knaus et~al.(1995)Knaus, Harrell, Lynn, Goldman, Phillips, Connors,
  Dawson, Fulkerson, Califf, Desbiens, et~al.]{knaus1995support}
William~A Knaus, Frank~E Harrell, Joanne Lynn, Lee Goldman, Russell~S Phillips,
  Alfred~F Connors, Neal~V Dawson, William~J Fulkerson, Robert~M Califf, Norman
  Desbiens, et~al.
\newblock The support prognostic model: objective estimates of survival for
  seriously ill hospitalized adults.
\newblock \emph{Annals of internal medicine}, 122\penalty0 (3):\penalty0
  191--203, 1995.

\bibitem[Linderman and Adams(2014)]{linderman2014discovering}
Scott Linderman and Ryan Adams.
\newblock Discovering latent network structure in point process data.
\newblock In \emph{International Conference on Machine Learning}, pages
  1413--1421, 2014.

\bibitem[Nagpal et~al.(2019)Nagpal, Sangave, Chahar, Shah, Dubrawski, and
  Raj]{nagpal2019nonlinear}
Chirag Nagpal, Rohan Sangave, Amit Chahar, Parth Shah, Artur Dubrawski, and
  Bhiksha Raj.
\newblock Nonlinear semi-parametric models for survival analysis.
\newblock \emph{arXiv preprint arXiv:1905.05865}, 2019.

\bibitem[Quattoni et~al.(2007)Quattoni, Wang, Morency, Collins, and
  Darrell]{quattoni2007hidden}
Ariadna Quattoni, Sybor Wang, Louis-Philippe Morency, Michael Collins, and
  Trevor Darrell.
\newblock Hidden conditional random fields.
\newblock \emph{IEEE Transactions on Pattern Analysis \& Machine Intelligence},
  pages 1848--1852, 2007.

\bibitem[Rosen and Tanner(1999)]{rosen1999mixtures}
Ori Rosen and Martin Tanner.
\newblock Mixtures of proportional hazards regression models.
\newblock \emph{Statistics in Medicine}, 18\penalty0 (9):\penalty0 1119--1131,
  1999.

\bibitem[Ruder(2016)]{ruder2016overview}
Sebastian Ruder.
\newblock An overview of gradient descent optimization algorithms.
\newblock \emph{arXiv preprint arXiv:1609.04747}, 2016.

\bibitem[Samo and Roberts(2015)]{samo2015scalable}
Yves-Laurent~Kom Samo and Stephen Roberts.
\newblock Scalable nonparametric bayesian inference on point processes with
  gaussian processes.
\newblock In \emph{International Conference on Machine Learning}, pages
  2227--2236, 2015.

\bibitem[Schumacher et~al.(1994)Schumacher, Bastert, Bojar, Huebner,
  Olschewski, Sauerbrei, Schmoor, Beyerle, Neumann, and
  Rauschecker]{schumacher1994randomized}
M~Schumacher, G~Bastert, H~Bojar, K~Huebner, M~Olschewski, W~Sauerbrei,
  C~Schmoor, C~Beyerle, RL~Neumann, and HF~Rauschecker.
\newblock Randomized 2 x 2 trial evaluating hormonal treatment and the duration
  of chemotherapy in node-positive breast cancer patients. german breast cancer
  study group.
\newblock \emph{Journal of Clinical Oncology}, 12\penalty0 (10):\penalty0
  2086--2093, 1994.

\bibitem[Simic(2009)]{simic2009upper}
Slavko Simic.
\newblock On an upper bound for jensen’s inequality.
\newblock \emph{Journal of Inequalities in Pure and Applied Mathematics}, 2009.

\bibitem[Snoek et~al.(2013)Snoek, Zemel, and Adams]{snoek2013determinantal}
Jasper Snoek, Richard Zemel, and Ryan~P Adams.
\newblock A determinantal point process latent variable model for inhibition in
  neural spiking data.
\newblock In \emph{Advances in Neural Information Processing Systems}, pages
  1932--1940, 2013.

\bibitem[Sutton et~al.(2012)Sutton, McCallum, et~al.]{sutton2012introduction}
Charles Sutton, Andrew McCallum, et~al.
\newblock An introduction to conditional random fields.
\newblock \emph{Foundations and Trends{\textregistered} in Machine Learning},
  4\penalty0 (4):\penalty0 267--373, 2012.

\bibitem[Yao(2014)]{yao2014methods}
Cindy~Q Yao.
\newblock \emph{Methods for the Identification of Biomarkers in Prostate and
  Breast Cancer}.
\newblock PhD thesis, University of Toronto (Canada), 2014.

\end{thebibliography}

\newpage

\appendix
\section{Appendix}
\subsection{Bounding Jensen's Inequality [Dragomir -- Loose Bound]}
\label{appendix:loose_bound}
\citep{simic2009upper} propose multiple distribution agnostic upper bounds for Jensen's inequality in the case of generic continuous convex functions defined on a compact set $[\mathcal{L}, \mathcal{U}]$. One of the popular bounds in this regime is the Dragomir's inequality proposed in \citep{dragomir1999converse}. Given $z_i$ is bounded with probability $\delta$, section \ref{sec:analysis} bounds $\rva, \rvb$ with $(\rva \in [\lrva, \urva], \rvb \in [\lrvb, \urvb])$. By Dragomir's \citep{dragomir1999converse} inequality for a convex function $f$,
\begin{align}
    \label{eq:dragomir}
    \E[f(\rva)] - f(\E(\rva)) \leq \frac{1}{4}(\urva-\lrva)(f'(\urva)-f'(\lrva))
\end{align}
%Sections \ref{appendix:loose_bound}, \ref{appendix} focus on bounding the gap for $\rva$ and the function $x^{-p}$. The proof for $\rvb$ is on similar lines. For $f(x)=x^p$ and $f(x)=x^{-p}$ \ref{eq:dragomir} would lead to the following bounds respectively,
For $f \in \{f_p\}_{p=2}^{\infty}, f_p(x)=x^p$ and $f \in \{f_p\}_{p=2}^{\infty}, f_p(x)=x^{-p}$, the bounds are given by eq. \ref{eq:dragomir_closed_form_1} and eq. \ref{eq:dragomir_closed_form_2} respectively.
\begin{align}
    \label{eq:dragomir_closed_form_1}
    \frac{p}{4}(\urvb-\lrvb)(\urvb^{p-1}-\lrvb^{p-1}) \\
    \label{eq:dragomir_closed_form_2}
    \frac{p}{4}(\urva-\lrva)(\lrva^{-(p+1)}-\urva^{-(p+1)})
\end{align}

This bound is easy to compute and can be visualized via the gap shown in figure \ref{fig:vis_bound}. This is also quite naive and generic since it only uses the first order conditions for convex functions to arrive at an upper bound. In the following section we provide a tighter upper bound under the stronger assumptions of strict convexity.

\subsection{Bounding Jensen's Inequality [Convex Conjugate -- Tight Bound]}
\label{appendix:tight_bound}
This section provides the proof for theorem \ref{theorem:main} in the main paper. We investigate bounds under the special case of strictly convex functions $\{\phi_p\}_{|p|=2}^{\infty}$ with $\phi_p : \R_{++} \to \R_{++}$ and $\phi_p(x)=x^p$. We also show  visually (figure \ref{fig:vis_bound}) that our bound is the tightest possible distribution agnostic bound for the given set of functions. With $\mathbb{G}(\rva, \phi_p)$ as the bound of interest,
\begin{align}
    \label{eq:lambda_max_formulation}
        \mathbb{G}(\rva, \phi_p) \leq \max_{\kappa \in [0,1]} \kappa \phi_p(\urva) + (1-\kappa) \phi_p(\lrva) - \phi_p(\kappa \urva + (1-\kappa) \lrva)
\end{align}
Figure \ref{fig:vis_bound} depicts geometrically the maximization problem in the RHS of eq. \ref{eq:lambda_max_formulation} which we solve via the convex conjugate of $\phi_p$. The optimization problem in eq. \ref{eq:lambda_max_formulation} involves identifying $\rva \in [\lrva, \urva]$, such that the line in figure $\ref{fig:vis_bound}$ denoted by $y=\phi'(\rva^*)\rva+c$ where $\rva^*$ is farthest away from $\phi(\rva)$ at the optimal point. 

Given that $\phi_p(x)=(1/x^p)$ is a closed proper strict convex function on $[\lrva, \urva]$, (the case when we consider $\phi_p(x)=x^p$ on $[\lrvb, \urvb]$ is similar) we can define the convex conjugate of $\phi_p(x)$. The form in eq. \ref{eq:conj_basics} is similar to what we need (distance between $\phi_p(x)$ and a line with slope defined by $\frac{\phi_p(\urva)-\phi_p(\lrva)}{\urva-\lrva}$

\begin{align}
    \label{eq:conj_basics}
    \phi_p^*(y) = \sup\limits_{x} yx-\phi_p(x) \quad  \partial \phi_p^*(y) = \{x : yx-\phi_p(x)=\phi_p^*(y)\}
\end{align}
\begin{align}
\label{eq:conj}
\partial \phi_p^*\left(\frac{\phi_p(\urva)-\phi_p(\lrva)}{\urva-\lrva}\right) = \argmax_{w \in [\lrva, \urva]} w . \frac{\phi_p(\urva)-\phi_p(\lrva)}{\urva-\lrva} - \phi_p(w). 
\end{align}
Since $(1/x^p)$ is strictly convex the subgradient of $\phi_p^*$ at $y$ is a singleton set and is exactly equal to the $\partial \phi_p^{*}(y)$, giving us a unique maximizer $w^*$ in eq. \ref{eq:conj}.
\begin{align}
\label{eq:conj_2}
\partial\phi_p^{*}(y) = \left(\frac{-p}{y}\right)^{\frac{1}{p+1}}  \quad   w^*= \left(p.\frac{\urva-\lrva}{\lrva^{-p}-\urva^{-p}}\right)^{\frac{1}{p+1}}
\end{align}
Eq. \ref{eq:conj_2} tells us that $w^*$ (and thus the bound in Eq. \ref{eq:lambda_max_formulation}) for a given sample $(x_i, t_i, R(t_i))$ is a function of $\beta, x_i$ \& $x_k \in R(t_i)$. 
Given $w*$ it is easy to compute $\mathbb{G}(\rva, \phi_p)$ by,

\begin{align}
\label{eq:uniq_min}
w^* = \kappa^* \urva + (1-\kappa^*) \lrva
\end{align}
\begin{align}
\kappa^* = \frac{w^*-\lrva}{\urva-\lrva}
\end{align}
\begin{align}
\mathbb{G}(\rva, \phi_p) = \kappa^* \phi_p(\urva) &+ (1-\kappa^*) \phi_p(\lrva) - \phi_p(\kappa^* \urva + (1-\kappa^*) \lrva)
\end{align}
In section \ref{sec:results} we empirically compared this bound with eqs. \ref{eq:dragomir_closed_form_1}, \ref{eq:dragomir_closed_form_2} from appendix \ref{appendix:loose_bound}.

\end{document}